\documentclass[sigconf]{acmart}
\AtBeginDocument{%
  \providecommand\BibTeX{{%
    \normalfont B\kern-0.5em{\scshape i\kern-0.25em b}\kern-0.8em\TeX}}}

\setcopyright{acmcopyright}
\copyrightyear{2023}
\acmYear{2023}
\acmDOI{XXXXXXX.XXXXXXX}

\acmConference[VAT@ ACM/IEEE HRI '23]{Workshop on Variable Autonomy in Human-Robot Teams @ ACM/IEEE HRI}{March 13--16,
  2023}{Stockholm, SE}
\acmPrice{15.00}
\acmISBN{978-1-4503-XXXX-X/18/06}

\usepackage{amsmath}
\usepackage{listings}
\usepackage{mathtools}

\usepackage{arydshln}

\usepackage[linesnumbered,ruled]{algorithm2e}
\usepackage{algorithmic}

\usepackage{tikz}

\usetikzlibrary{trees}
\usetikzlibrary{shapes}
\usetikzlibrary{positioning}
\usetikzlibrary{automata, arrows}
\usetikzlibrary{calc}
\usetikzlibrary{fit}
\tikzstyle{vertex}=[circle,fill=black!35,minimum size=30pt,inner sep=0pt,align=center]
\tikzstyle{activity}=[circle,fill=green!35,minimum size=5pt,inner sep=0pt,align=center, font=\small]
\tikzstyle{heuristic}=[circle, fill=black!10, minimum size=8pt,inner sep=0pt, font=\scriptsize]
\tikzstyle{selected vertex} = [vertex, fill=red!24]
\tikzstyle{nn node} = [vertex, minimum size=8pt, align=center, fill=red!24]
\tikzstyle{small nn node} = [vertex, fill=black!35, minimum size=2pt, align=center, fill=red!24]
\tikzstyle{start vertex} = [vertex, fill=green!30]
\tikzstyle{goal vertex} = [vertex, fill=blue!40]
\tikzstyle{edge} = [draw,thick,-stealth,line width=0.5pt]
\tikzstyle{weight} = [font=\small]
\tikzstyle{selected edge} = [draw,line width=5pt,-,red!50]
\tikzstyle{ignored edge} = [draw,line width=5pt,-,black!20]
\tikzstyle{place}= [circle,thick,draw=black,fill=blue!50,minimum size=6mm]
\tikzstyle{trans}=[rectangle,thick,fill=black,minimum width=6mm,inner ysep=2pt]
\tikzstyle{ctl node} = [vertex, minimum size=15pt, %font=\small,
align=center, fill=red!24]
\tikzstyle{model-node} = [vertex, minimum size=10pt, align=center, fill=red!24]

\tikzstyle{box}=[minimum height=.5cm,minimum width=4cm,fill=white,opacity=.9,rounded corners,thin]
\tikzstyle{partition}=[minimum height=.5cm,minimum width=1cm,fill=white,opacity=.9,thin]
\tikzstyle{exec node} = [box, minimum size=10pt, align=center, fill=yellow!90]
\tikzstyle{cond node} = [exec node, fill=green!50!black]
\tikzstyle{deco node} = [diamond, exec node, fill=purple!70]

\usepackage[nolist]{acronym}

\begin{acronym}
    \acro{adl}[ADL]{Activity of Daily Living}
    \acroplural{adl}[ADLs]{Activities of Daily Living}
    \acro{sct}[SCT]{Shared Control Template}
    \acroplural{sct}[SCTs]{Shared Control Templates}
    \acro{phast}[PHAST BT]{PHAse Switching Teleoperation Behavior Tree}
    \acroplural{phast}[PHAST BTs]{PHAse Switching Teleoperation Behavior Trees}
    \acro{bt}[BT]{Behavior Tree}
    \acroplural{bt}[BTs]{Behavior Trees}
    \acro{bci}[BCI]{Brain Computer Interface}
    \acro{dof}[DoF]{Degree of Freedom}
    \acroplural{dof}[DoFs]{Degrees of Freedom}
    \acro{lts}[LTS]{Labelled Transition System}
    \acroplural{lts}[LTS]{Labelled Transition Systems}
    \acro{ee}[EE]{End Effector}
    \acro{im}[IM]{Input Mapping}
    \acroplural{im}[IMs]{Input Mappings}
    \acro{ac}[AC]{Active Constraint}
    \acroplural{ac}[ACs]{Active Constraints}
    \acro{scan}[SCAN]{Shared Control Action Node}
    \acroplural{scan}[SCANs]{Shared Control Action Nodes}
    \acro{fsm}[FSM]{Finite State Machine}
    \acroplural{fsm}[FSMs]{Finite State Machines}
    
    \acro{lfd}[LfD]{Learning from Demonstration}
    
\end{acronym}
\newcommand{\treename}[0]{\ac{phast}}
\newcommand{\treenames}[0]{\acp{phast}}
\begin{document}

\title[Teleoperation using PHAST BTs]{Assistive Robot Teleoperation Using Behavior Trees}

\author{Mohamed Behery}
\orcid{0000-0003-1331-9419}
\affiliation{%
  \institution{Knowledge-Based Systems Group\\ RWTH Aachen University}
  \city{Aachen}
  \country{Germany}
  \postcode{52062}
}
\email{behery@kbsg.rwth-aachen.de}

\author{Minh Trinh}
\affiliation{%
  \institution{Laboratory of Machine Tools\\and Production Engineering\\RWTH Aachen University}
  \streetaddress{}
  \city{Aachen}
  \country{Germany}}
\email{m.trinh@wzl.rwth-aachen.de}

\author{Christian Brecher}
\affiliation{%
  \institution{Laboratory of Machine Tools\\and Production Engineering\\RWTH Aachen University}
  \city{Aachen}
  \country{Germany}
}
\email{c.brecher@wzl.rwth-aachen.de}

\author{Gerhard Lakemeyer}
\affiliation{%
  \institution{Knowledge-Based Systems Group\\ RWTH Aachen University}
  % \streetaddress{P.O. Box 1212}
  \city{Aachen}
  \country{Germany}
  \postcode{52062}
}
\email{gerhard@kbsg.rwth-aachen.de}

\renewcommand{\shortauthors}{Behery, et al.}

\begin{abstract}
Robotic assistance in robot arm teleoperation tasks has recently gained a lot of traction in industrial and domestic environments.
A wide variety of input devices is used in such setups.
Due to the noise in the input signals (e.g., \acl{bci}) or delays due to environmental conditions (e.g., space robot teleoperation), users need assistive autonomy that keeps them in control while following predefined trajectories and avoiding obstacles.
This assistance calls for activity representations that are easy to define by the operator and able to take the dynamic world state into consideration.
This paper represents \aclp{adl} using \acp{bt} whose inherent readability and modularity enables an end user to define new activities using a simple interface.
To achieve this, we augment \acp{bt} with \aclp{scan}, which guide the user's input on a trajectory facilitating and ensuring task execution. %performance.
\end{abstract}





% \received{30 January 2023}

\maketitle
\acresetall
\section{Introduction}
\label{sec:intro}
Recent work has paved the way for robot arm teleoperation in industrial and domestic domains.
Use cases for robot teleoperation can be seen industrial metal forging~\cite{behery2020action}, robot-assisted surgery~\cite{enayati2016haptics}, and \ac{adl} execution by people with tetraplegia~\cite{behery2016knowledge,Quere2020shared} to name a few.
All these cases emphasise the need for shared control autonomy where a smart agent assists the human operator keep the robot on a predefined trajectory, avoiding obstacles, and achieving the goal of the given task.
Since some of these use cases heavily rely on user experience, they cannot be replaced by fully autonomous solutions as they deprive the user of flexibility and full control over the activity (e.g., robot assisted surgery).
However, input devices vary according to the use case and can have varying numbers of degrees of freedom and noise levels.
Extracting user input typically requires translating the low dimensional input from the input device into the high \acp{dof} robot workspace \cite{Quere2020shared,behery2016knowledge}. 
Many approaches have recently tackled this issue by embedding an action representation within the robot control framework \cite{Quere2020shared,behery2016knowledge,muelling2017autonomy}, however, most approaches focus on the usability of teleoperation, the intuitiveness of the robot trajectory \cite{luo2021trajectory}, seamlessness of the transition between direct control and shared autonomy \cite{bustamante2021toward}, as well as integration with whole body control \cite{Quere2020shared,luo2020teleoperation}.
These approaches do not handle the extensibility neither when creating new actions nor when modifying existing ones. 

Recent action description approaches focus on human interface when programming a robot (or intelligent agent). 
The use of \acp{bt} is becoming a popular action description approach because of their usability and modularity allowing users to create, modify, and maintain task definitions on a high level. 
A \ac{bt} \cite{colledanchise2018behavior} is a model representing the behavior of an agent or a robot. 
They are easily maintainable compared to other approaches such as \acp{fsm}.
As shown in \cite{olsson2016behavior}, they have a higher maintainability index \cite{oman1992metrics,coleman1994using} compared to \acp{fsm} that implement the same tasks.
Maintainability index is measured using several software characteristics including maintenance intensity, reuse, and development effort.

In this paper, we propose a new framework for robot teleoperation allowing users to define actions using \acp{bt}.
We introduce \treenames{}, which are a structurally restricted class of \acp{bt} acting as activity templates focusing on teleoperation for execution and modeling of \acp{adl}, inspired by other approaches \cite{Quere2020shared,behery2016knowledge} that show great promise in this task.
The proposed \treenames{} are an improvement to the state of the art by virtue of the modularity, extensibility, and readability they inherit from \acp{bt}.
These \ac{bt} properties have been discussed extensively in numerous works \cite{colledanchise2018behavior,iovino2020survey,paxton2017costar}.

\begin{figure}[t]
    \vspace{6pt}
    \centering
    \includegraphics[width=0.95\columnwidth]{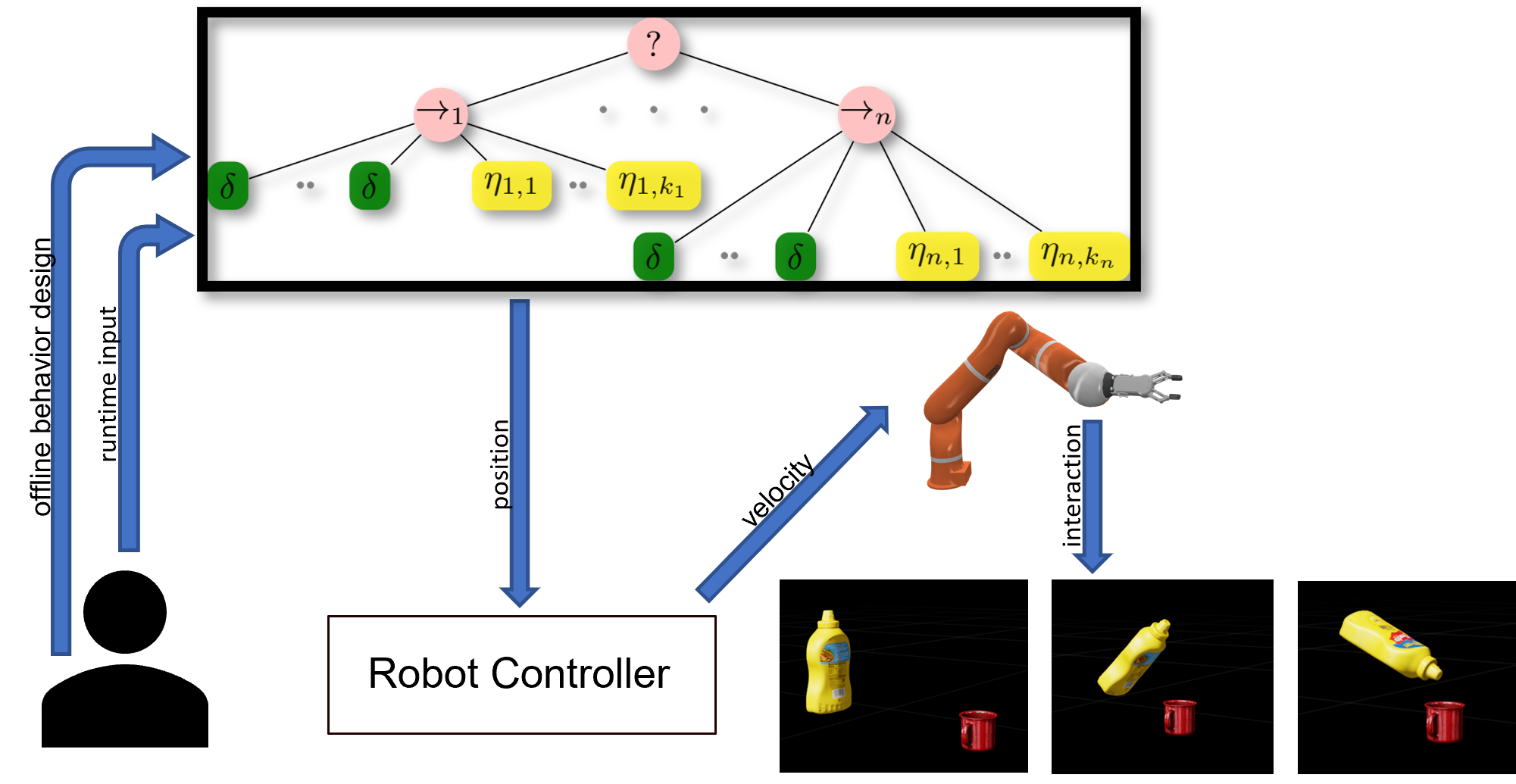}
    \caption{An overview of the proposed approach where a user can describe an activity as a \ac{bt} offline.
    The \ac{bt} is then used along with the user's input to assist with the teleoperation of the robot in real time by sending the modified position commands to the robot controller that can transform them into velocity commands for interacting with the environment.}
    \label{fig:overview}
\end{figure}

Figure~\ref{fig:overview} shows an overview of our approach. 
We allow a user to describe an \ac{adl} as a \treename{} which is then executed by the robot to guide the user's commands to a predefined trajectory while executing tasks.
The readability and structure of \acp{bt} enable users to design new trees as well as modify existing trees or using them together to compose new trees through a simple user interface.
For example, drag-and-drop interfaces are developed in \cite{olsson2016behavior,paxton2017costar} for building and visualization of \acp{bt} in addition to other tools. 

This novel application domain for \acp{bt} opens the door to industrial applications of smart robotic teleoperative assistance in use cases studied in \cite{behery2020action,baier2022framwork}.
We plan to empirically verify this work by conducting user studies allowing users to define activities and teleoperate the robot using the trees in domestic assistance use cases.
Additionally, we will conduct experiments allowing domain experts of different industrial teleoeperation scenarios to define activities and teleoperate robots in industrial use cases.

This paper is structured as follows: 
The state of the art approaches on robot teleoperation for \ac{adl} execution are discussed in Section~\ref{sec:sota}. 
Next we give an overview on \acp{bt} in Section~\ref{sec:background}. 
Section~\ref{sec:phast-bt} describes our proposed approach to represent \acp{adl} using \acp{bt}.
To that end, we introduce new node types described in Section~\ref{sec:scan}. 
Next, we discuss a use case of shared autonomy teleoperation in Section~\ref{sec:use-case}.
Finally, Section~\ref{sec:conc} concludes with an outlook on future work.

\section{Background and Related Work}
\label{sec:sota}

Assistive robotics gained a lot of traction in offering domestic assistance for the physically impaired. 
Recent studies have shown that the use of shared autonomy reduces the physical and mental workload involved in robot teleoperation~\cite{lin2020shared} resulting in a preference towards shared autonomy in performing \acp{adl}.
However, most approaches focus on the usability during task performance as well as the integration with the Whole Body Control~\cite{Quere2020shared,behery2016knowledge,luo2020teleoperation} and switching between shared autonomy and direct user control~\cite{bustamante2021toward}.
The focus on usability is restricted on reducing the mental effort of teleoperation either by using haptic feedback~\cite{zhang2021haptic} or using motion prediction to generate more intuitive trajectories~\cite{luo2021trajectory}.

Extending the number of supported actions using these approaches requires programming and robotics experience in addition to time and effort.
This motivated \ac{lfd} approaches such as~\cite{quere2021learning} which uses \ac{lfd} to teach motion primitives to a robot. 
However, this approach builds on \acp{sct}~\cite{Quere2020shared} which still suffer from the short comings of \acp{fsm} and \acp{lts} as a high level control language, e.g., the number of connections grows exponentially in the number of states.
Other approaches for task description exist \cite{beetz2013cognition,tenorth2009knowrob} but require knowledge from an underlying ontology \cite{paxton2017costar}.

This is why we propose exploiting the readability, modularity, and composability of \acp{bt} for this task.
They have shown great success in allowing people with no programming experience to program robotic agents in computer games~\cite{gladiabots}.
\acp{bt} have only been used alongside teleoperation in \cite{hu2015semi} to control a robot in semi-autonomous surgery.
However, they do not use the shared autonomy control paradigm which has recently shown great success~\cite{behery2016knowledge,Quere2020shared,muelling2017autonomy}.
In this paper we introduce shared autonomy control as a use case of \acp{bt}.

In our approach, we build on \acp{bt} exploiting many of their favorable properties giving users the ability to define tasks with no programming experience.
In the next section, we describe how a \ac{bt} is used to model robotic tasks.

\subsection{Behavior Trees}
\label{sec:background}
In this section we overview \acp{bt} as a high-level robot control structure.
We also describe their inner workings and structure which give them composability, modularity, and readability compared to other approaches.

A \ac{bt} is a tick-based modular behavior model with a tree structure. A \emph{tick} is a signal that triggers the execution of a node in the tree.
When the root node is ticked it signals the execution of the whole tree and propagates the tick through its children to the leaves. 
Each node handles the tick depending on its type and returns its status to its parent (except for the root).
Control flow is achieved through internal nodes that control which of their children to tick.
They are classified into \emph{Sequence}, which execute their respective children in sequence and \emph{Fallback} nodes whose children define alternative execution paths\footnote{\acp{bt} also offer parallel execution nodes, but these are not used in this paper.}.
\acp{bt} also provide \emph{Decorator} nodes for custom control flow behavior (e.g., repeat $n$ times).
Additionally, the leaf-node types are split into \emph{Action} and \emph{Condition} nodes.
As condition and fallback nodes are used in combination to define specific conditions that must be met in order to execute an action and alternative actions otherwise, the \ac{bt} becomes reactive to plan and world state changes.
\acp{bt} are similar to \acp{fsm}, except that they consist of tasks rather than states.
\acp{fsm} possess poor scalability, as the addition of new states is not trivial \cite{colledanchise2018behavior}. 
Compared to BTs, extending or reusing \acp{fsm} may get extremely challenging. 
Due to the simple tree structure, \acp{bt} can help robot programmers raise the abstraction level for simpler implementation of robot tasks in higher-level representations.
Additionally, editors such as~\cite{paxton2017costar,gladiabots} show that \acp{bt} are easy to understand and design task plans for a variety of use-cases by users of varying programming and robotics experience.
They have also been extended and used in variety of use cases and robotics applications\footnote{We do not discuss these here due to space limitations. However, ~\cite{iovino2020survey} surveys these extensions and applications.}.
The next section describes how we exploit the properties of \ac{bt} to allow end users to define robot teleoperation tasks with little to no programming experience.

\section{Phase Switching Teleoperation Behavior Trees}
\label{sec:phast-bt}

\treenames{} are a structurally restricted class of \acp{bt}, we limit the structure of the tree to reduce user decisions and thus the mental effort required to design the tree and introduce the \ac{scan} which modifies the user input during operation.
This input modification has shown great success in earlier approaches~\cite{behery2016knowledge,Quere2020shared,muelling2017autonomy}, which lacked extensibility.

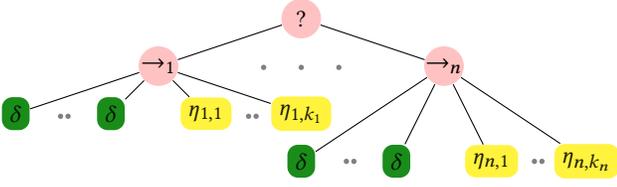
\begin{figure}
    \vspace{6pt}
	\centering
	\begin{tikzpicture}[scale=1,level distance=2em,
	level 1/.append style={sibling distance=12em},
	level 2/.append style={sibling distance=4em},]
		\node [ctl node] {$?$}
				child {
    				node [ctl node] {$\rightarrow_{1}$}
        				child {node [cond node] {$\delta$}}
        				child {node [cond node] {$\delta$}}
    				child {node [exec node] {$\eta_{1, 1}$}}
    				child {node [exec node] {$\eta_{1, k_{1}}$}}}
				child {
    				node [ctl node] {$\rightarrow_{n}$} [level distance=4em]
        				child {node [cond node] {$\delta$}}
        				child {node [cond node] {$\delta$}}
    				child {node [exec node] {$\eta_{n, 1}$}}
    				child {node [exec node] {$\eta_{n, k_{n}}$}}};
    				
		\node[nn node,color=black!50,minimum size=2pt] () at  (0.5,-0.65)   	{}; 
		\node[nn node,color=black!50,minimum size=2pt] () at  (0.0,-0.65)   	{}; 
		\node[nn node,color=black!50,minimum size=2pt] () at  (-0.5,-0.65)   	{}; 
		
		\node[nn node,color=black!50,minimum size=2pt] () at  (-3.1,-1.3)   	{}; 
		\node[nn node,color=black!50,minimum size=2pt] () at  (-3.2,-1.3)   	{}; 
		
		\node[nn node,color=black!50,minimum size=2pt] () at  (-0.7,-1.3)   	{}; 
		\node[nn node,color=black!50,minimum size=2pt] () at  (-0.6,-1.3)   	{}; 
		
		\node[nn node,color=black!50,minimum size=2pt] () at  (3.1,-1.9)   	{}; 
		\node[nn node,color=black!50,minimum size=2pt] () at  (3.2,-1.9)   	{}; 
		
		\node[nn node,color=black!50,minimum size=2pt] () at  (0.6,-1.9)   	{}; 
		\node[nn node,color=black!50,minimum size=2pt] () at  (0.7,-1.9)   	{}; 
	\end{tikzpicture}
	\caption{Example \treename{}.
	Limited to three levels and a fallback node for the root. 
        The second layer contains only sequence nodes, each corresponds to an activity phase. 
        Their children are condition nodes for controlling the activation of \acp{scan} of different phases.	
         }
	\label{fig:sctt-structure}
\end{figure}

The general structure of the \treename{} is shown in Figure~\ref{fig:sctt-structure}.
Similar to previous approaches~\cite{behery2016knowledge,Quere2020shared}, we describe an activity as a list of phases, each of which is represented as a sequence node on the second level of the tree.
We limit the \acp{bt} to three levels with a fallback root node which is ticked with every user input. 
The leaf nodes are either condition nodes to control phase activation or \acp{scan} that perform the input modification.

\subsection{Shared Control Action Nodes}
\label{sec:scan}
Our proposed \acp{scan} work similar to the \acp{im} and the \acp{ac} used in the \acp{sct} \cite{Quere2020shared}.
\treenames{} provide \acp{scan} that encapsulate auxiliary functions used to modify the user's input to perform simple mappings and actively constrain the \ac{ee} movement, respectively.
These nodes first calculate the next \ac{ee} pose given the current one and the user's input.
For the remainder of this paper, we will use $f: \mathbb{R}^{3} \times \mathbb{R}^{6} \rightarrow \mathbb{R}^{6}$ to refer to an auxiliary function used in \acp{scan} that takes the user input $u\in \mathbb{R}^{3}$ and a pose $e\in \mathbb{R}^{6}$ and maps them to the resulting new \ac{ee} pose $e^{\prime}\in\mathbb{R}^{6}$.

\acp{scan} can read the \ac{ee} pose from a world state, calculate the next pose using the respective predefined function, and send it to the low level controller.
Equation~\ref{eq:scan-tick} shows the tick function of the \ac{scan}, the function $f$ returns the next state $s_{t+1}$, e.g., \ac{ee} pose, given the current one $s_{t}$ and the user input $u$. 

\begin{equation}
    \label{eq:scan-tick}
    s_{t+1} = f(u, s_{t})
\end{equation}

The return status of these nodes is always TRUE so they don't need to return RUNNING since they update the \ac{ee} pose in small steps inversely proportional to the tick frequency of the tree. 

\section{Case Study}
\label{sec:use-case}

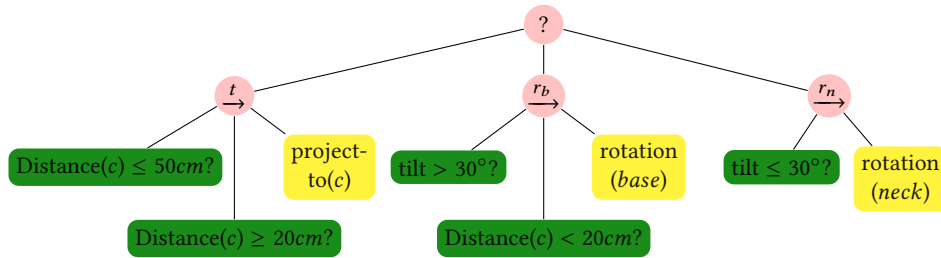
\begin{figure*}[ht]
    \vspace{6pt}
	\centering
	\begin{tikzpicture}[scale=1,level distance=3em,
	level 1/.append style={sibling distance=12em},
	level 2/.append style={sibling distance=4em},]
		\node [ctl node] {$?$}
            child [sibling distance=13em]{
                node [ctl node] {$\xrightarrow{t}$}
                child [sibling distance=5em] {node [cond node] {Distance($c$) $\le 50cm?$}}
                child [level distance=6em, sibling distance=2em] {node [cond node] {Distance($c$) $\ge 20cm?$}}
                % child [level distance=3.5em] {node [exec node] {horizontal\\motion()}}
                child [level distance=3em, sibling distance=4em] {node [exec node] {project-\\to($c$)}}}
            child {
                node [ctl node] {$\xrightarrow{r_{b}}$}
                child [sibling distance=4em] {node [cond node] {tilt $> 30^{\circ}?$}}
                child [level distance=6em, sibling distance=5em] {node [cond node] {Distance($c$) $< 20cm?$}}
                child [sibling distance=4em] {node [exec node] {rotation\\($base$)}}}
            child {
                node [ctl node] {$\xrightarrow{r_{n}}$}
                child {node [cond node]{tilt $\le 30^{\circ}?$}}
                child [level distance=3.5em, sibling distance=6em] {node [exec node] {rotation\\($neck$)}}};
	\end{tikzpicture}
	\caption{A \treename{} representing the activitiy of pouring water from a bottle.
 The activity has $3$ phases, translating the bottle $t$, rotating around the base $b$, and rotating around the neck $n$.}
	\label{fig:demo}
\end{figure*}

This section describes a use case for representing an \ac{adl} using \treenames{}.
We describe the activity of pouring liquid from a bottle $b$ to a cup $c$.
This activity was chosen as an example as it was used as an example in previous work~\cite{behery2016knowledge,Quere2020shared}.
Here we also give examples of the input mapping functions that can be used in the \acp{scan}.
The activity is divided into $3$ phases $t$, $r_{b}$, and $r_{n}$, as seen in Figure~\ref{fig:overview}.
Phase $t$ is the translation to place the bottle at an adequate distance from the cup.
In phases $r_{b}$, and $r_{n}$, we rotate the bottle around its base and neck, respectively.
These are represented in Figure~\ref{fig:demo} as the sequence nodes $\xrightarrow{t}$, $\xrightarrow{r_{b}}$, and $\xrightarrow{r_{n}}$, respectively.
The condition nodes measure the distance between the bottle and the cup as well as the bottle's tilt over the horizon.

In the first phase, the user input is projected on a straight line from the bottle to the cup, it starts when the bottle is $50$~cm away from the cup till $20$~cm and projects the input on the line between the bottle location $l_{b}$ and the cup location $l_{c}$.
When the user commands lead to a new location of the bottle (held in the gripper), $l^{u}_{b}$.
We interpolate the line between the bottle location $l_{b}$ and $l^{u}_{b}$ as $\vec{l^{u}_b}~=~l_{b}~-~l^{u}_b$.
Similarly, we interpolate a line between the locations of the bottle and cup, $\vec{L_{b,c}}~=~l_{b}~-~l_{c}$.
After that we project $l^{u}_{b}$ on $\vec{L_{b,c}}$ using $project-to$ function that performs Equation~\ref{eq:projection} to calculate the new location of the bottle $l^{\prime}_{b}$ and uses it as the translation component in the new pose sent to the robot controller.
\begin{equation}
    \label{eq:projection}
    l^{\prime}_{b}~ =~l_{b}~+~(\vec{l^{u}_b}~\cdot~||\vec{L_{b,c}}||)~*~||\vec{L_{b,c}}||)
\end{equation}

In the second and third phases, the rotation function seen in Equation~\ref{eq:rotation-axis} that rotates the given object around a given point on the bottle.
We calculate the axis of rotation $r$ as the cross product shown in Equation~\ref{eq:rotation-axis}, where $\vec{L_{1}} = l_{c} - p_{c}$ is the line between the cup's location and it's pouring point $p_{c}$ and $\vec{L_{2}} = l_{b} - p_{c}$ is the line between the bottle's location and $p_{c}$.
We scale the user's input to get the rotation angle $\theta$.
This is done to tie the rotation speed and direction with the user's input allowing them to cancel the operation by giving commands in the opposite direction.
The rotation matrix is shown in Equation~\ref{eq:rotation-matrix}, where $c$ and $s$ are $cos(\theta)$ and $sin(\theta)$ respectively.

\begin{equation}
    \label{eq:rotation-axis}
    \Vec{r} = \begin{bmatrix} x \\ y\\ z \end{bmatrix}~=~L_{1}~\times~L_{2}
\end{equation}
\begin{equation}
    \label{eq:rotation-matrix}
            \begin{bmatrix} 
                c + x^{2}(1-c) & xy(1-c) - zs       & xz(1-c) + ys & 0 \\
                xy(1-c) + zs & c + y^{2}(1-c)       & yz(1-c) - zs & 0 \\
                xz(1-c) - ys & yz(1-c) + xs       & c + z^{2}(1-c) & 0\\
                0 & 0 & 0 & 1
            \end{bmatrix}
\end{equation}
The rotation function used in the \acp{scan} in Figure~\ref{fig:demo} uses this rotation matrix for calculating the pose of the bottle in the next step.

Although these auxiliary functions seem rather complex to program, they can easily be encapsulated in predefined nodes as seen in several \ac{bt} editors~\cite{paxton2017costar,gladiabots}.
They offer composable predefined leaf nodes to create different behaviors using drag-and-drop interface.

\section{CONCLUSION AND OUTLOOK}
\label{sec:conc}
This paper introduces \treenames{}, a high level robot control model that results from applying some structural restrictions on \acp{bt}.
We show how \treenames{} can be used for shared autonomy robot teleoperation, a novel application for \acp{bt}.
\treenames{} allow a user to describe an activity and execute it using shared autonomy, which maintain the user satisfaction and usability by exploiting the modularity, composability, and extensibility of traditional \acp{bt}.
At the same time, we can maintain the ability to introspect the tree as well as the trust and predictability of the robot's actions by having a live display interface similar to the one mentioned in \cite{Quere2020shared} displaying the states for each node in real time.
We plan to empirically verify this approach in the near future by conducting experiments examining both task description and execution.

%%%%%%%%%%%%%%%%%%%%%%%%%%%%%%%%%%%%%%%%%%%%%%%%%%%%%%%%%%%%%%%%%%%%%%%%%%%%%%%%

\section*{Acknowledgments}
We would like to thank the anonymous reviewers for their valuable feedback.
Funded by the Deutsche Forschungsgemeinschaft (DFG, German Research Foundation) under Germany's Excellence Strategy–EXC-2023 Internet of Production – 390621612. 
This work is partially supported by the EU ICT-48 2020 project TAILOR (No. 952215)
\bibliographystyle{IEEEtran}
\bibliography{main.bib}

\end{document}